\documentclass[runningheads]{llncs}
\usepackage{graphicx}

\usepackage{tikz}
\usepackage{comment}
\usepackage{amsmath} % define this before the line numbering.
\usepackage{amssymb}
\usepackage{color}
\usepackage{nicefrac}
\usepackage{adjustbox}
\usepackage[accsupp]{axessibility}  % Improves PDF readability for those with disabilities.

\usepackage{graphicx}
\usepackage{amsmath}
\usepackage{amssymb}
\usepackage{booktabs}
\usepackage{multirow}
\usepackage{tikz}  %%% change file in input in eso-pic.sty
\usepackage{pgfplots}
\usetikzlibrary{fit}
\pgfplotsset{compat=newest}%
\usepackage{enumitem}
\usepackage[pagebackref=true,breaklinks=true,colorlinks,bookmarks=false,citecolor=blue,linkcolor=blue]{hyperref}

\usepackage{subcaption}

\usepackage[width=122mm,left=12mm,paperwidth=146mm,height=193mm,top=12mm,paperheight=217mm]{geometry}

 \def\Eg{\emph{E.g}.}
\def\ie{\emph{i.e}.} 
 
\def\etc{\emph{etc}.} \def\vs{\emph{vs}.}

\def\etal{\emph{et al}.}

\begin{document}

\pagestyle{headings}
\mainmatter

\title{Video Instance Segmentation via Multi-scale Spatio-temporal Split Attention Transformer} 
\titlerunning{MS-STS Attention Transformer}
\author{Omkar Thawakar$^{1}$, Sanath Narayan$^{2}$, Jiale Cao$^{3}$ , Hisham Cholakkal$^{1}$, \\ Rao Muhammad Anwer$^{1}$, Muhammad Haris Khan$^{1}$, Salman Khan$^{1}$, \\ Michael Felsberg$^{4}$, Fahad Shahbaz Khan$^{1,4}$}
\authorrunning{O. Thawakar et al.}
\institute{$^1$Mohamed bin Zayed University of Artificial Intelligence, UAE \\
           $^2$Inception Institute of Artificial Intelligence, UAE \\
           $^3$Tianjin University, China \qquad $^4$Linköping University, Sweden}
\maketitle

\begin{abstract}
State-of-the-art transformer-based video instance segmentation (VIS) approaches typically utilize either single-scale spatio-temporal features or per-frame multi-scale features during the attention computations. We argue that such an attention computation ignores the multi-scale spatio-temporal feature relationships that are crucial to tackle target appearance deformations in videos. To address this issue, we propose a transformer-based VIS framework, named MS-STS VIS, that comprises a novel multi-scale spatio-temporal split (MS-STS) attention module in the encoder. The proposed MS-STS module effectively captures spatio-temporal feature relationships at multiple scales across frames in a video. We further introduce an attention block in the decoder to enhance the temporal consistency of the detected instances in different frames of a video. Moreover, an auxiliary discriminator is introduced during training to ensure better
foreground-background separability within the multi-scale spatio-temporal feature space. We conduct extensive experiments on two benchmarks: Youtube-VIS (2019 and 2021). Our MS-STS VIS achieves state-of-the-art performance on both benchmarks. When using the ResNet50 backbone, our MS-STS achieves a mask AP of 50.1\%, outperforming the best reported results in literature by 2.7\% and by 4.8\% at higher overlap threshold of AP$_{\mathtt{75}}$, while being comparable in model size and speed on Youtube-VIS 2019 val. set. When using the Swin Transformer backbone, MS-STS VIS achieves mask AP of 61.0\% on Youtube-VIS 2019 val. set. Our code and models are available at \url{https://github.com/OmkarThawakar/MSSTS-VIS}.
\end{abstract}

\section{Introduction}
Video instance segmentation (VIS) is a challenging computer vision problem with numerous real-world applications, including intelligent video analysis and autonomous driving. Given a video sequence, the task is to simultaneously segment and track all object instances from a set of semantic categories. The problem is particularly challenging since the target object needs to be accurately segmented and tracked despite appearance deformations due to several real-world issues such
as, target size variation, aspect-ratio change and fast motion. 

\begin{figure}[t!]
\centering
\resizebox{\textwidth}{!}{
\begin{minipage}{0.45\textwidth}
	\resizebox{6.0cm}{!}{%
	\includegraphics[width=\textwidth]{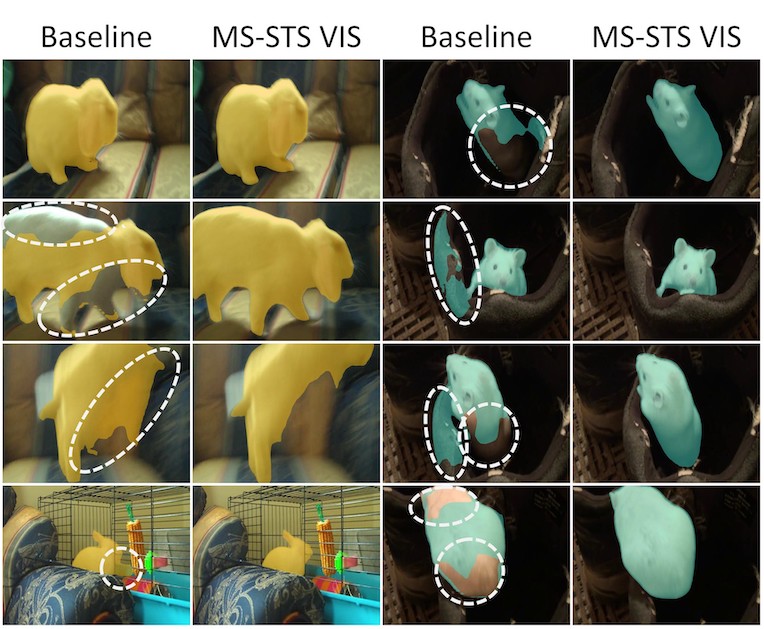}}
\end{minipage}
\qquad
\begin{minipage}{0.45\textwidth}
	\adjustbox{width=1\textwidth}{
	\begin{tikzpicture}
\begin{axis}[
axis lines = left,
ymin =35, ymax= 100,
xmin =30, xmax = 52,
xlabel= Accuracy (AP),
ylabel= Parameters (M),
]
\coordinate (legend) at (axis description cs:0.95,0.55);

\addplot[only marks,
mark=otimes*, purple,
mark size=3.5pt
]
coordinates {
(31.1, 58.1)};\label{plot:MaskTrack-RCNN}

\addplot[only marks,
mark=otimes*,  gray,
mark size=3.5pt
]
coordinates {
(30.6, 50.5)};\label{plot:STEmSeg}

\addplot[only marks,
mark=otimes*, orange,
mark size=3.5pt
]
coordinates {
(40.4, 69.0)};\label{plot:Propose-Reduce}

\addplot[only marks,
mark=otimes*, green,
mark size=3.5pt
]
coordinates {
(36.1, 40)};\label{plot:PCAN}

\addplot[only marks,
mark=otimes*, pink,
mark size=3.5pt
]
coordinates {
(34.8, 37.5)};\label{plot:Cross-VIS}

\addplot[only marks,
mark=otimes*, violet,
mark size=3.5pt
]
coordinates {
(36.2, 57.2)};\label{plot:VisTR}

\addplot[only marks,
mark=otimes*, blue,
mark size=3.5pt
]
coordinates {
(42.8, 40)};\label{plot:IFC}

\addplot[only marks,
mark=otimes*, cyan,
mark size=3.5pt
]
coordinates {
(47.4, 49.3)};\label{plot:SeqFormer}

\addplot[only marks,
mark=triangle*, magenta,
mark size=5pt
]
coordinates {
(50.1, 50.6)};\label{plot:Ours}

\end{axis}
\node[draw=none,fill=none, anchor= south east] at 
(legend){\adjustbox{width=0.96\textwidth}{
\begin{tabular}[ht!]{l|ccc|c}
\hline
Method & AP &AP$_{\mathtt{50}}$  &AP$_{\mathtt{75}}$ & Params(M) \\ \hline 

\ref{plot:MaskTrack-RCNN} MaskTrack-RCNN~\cite{MaskRCNN}  & 30.3 & 51.1 & 32.6 & 58.1\\
\ref{plot:STEmSeg} STEmSeg~\cite{STEmSeg}  & 30.6 & 50.7  & 33.5 & 50.5\\
\ref{plot:PCAN} PCAN~\cite{PCAN}  & 36.1 & 54.9  & 39.4 & 40.0\\
\ref{plot:Cross-VIS} Cross-VIS~\cite{CROSS-VIS}  & 34.8 & 56.8  & 38.0 & \textbf{37.5}\\  
\ref{plot:VisTR} VisTR~\cite{vistr}  & 36.2 & 59.8  & 36.9 & 57.2 \\ 
\ref{plot:Propose-Reduce} Propose-Reduce~\cite{propose_reduce}  & 40.4 &63.0  & 43.8 & 69.0\\
\ref{plot:IFC} IFC~\cite{ifc} &  42.8 & 65.8 &  46.8 & 39.3 \\ 
\ref{plot:SeqFormer} SeqFormer~\cite{SeqFormer}  & 47.4 & 69.8 & 51.8 & 49.3 \\ 
\ref{plot:Ours} \textbf{MS-STS VIS }  & \textbf{50.1} & \textbf{73.2}  & \textbf{56.6} & 50.6  \\  \hline
\end{tabular}  }};
\end{tikzpicture}
}
\end{minipage}

}\vspace{-0.1cm}
\caption{
\textbf{On the left:} Qualitative comparison of our MS-STS VIS with the baseline in the case of target appearance deformations on two example videos. For each method, four sample video frames are shown along the columns. The baseline struggles to accurately predict the \textit{rabbit} class instance undergoing appearance deformations due to scale variation, aspect-ratio change and fast motion. As a result, the predicted mask quality is hampered (marked by white dotted region). Similarly, the mask quality is also deteriorated for the \textit{mouse} class. Our MS-STS VIS addresses these issues by capturing multi-scale spatio-temporal feature relationships, leading to improved mask quality. Best viewed zoomed in. \textbf{On the right:} Accuracy (AP) \vs{} model size (params) comparison with existing methods using a single model and single-scale inference on YouTube-VIS 2019 val. set. We also report performance at AP$_{\mathtt{50}}$ and AP$_{\mathtt{75}}$. All methods here utilize the same ResNet50 backbone. Compared to the best reported results in literature~\cite{SeqFormer}, our MS-STS VIS achieves absolute gains of 2.7\% and 4.8\% in terms of overall mask AP and at higher overlap threshold of AP$_{\mathtt{75}}$, respectively, while being comparable in model size and speed (\cite{SeqFormer}: 11 FPS \vs{} \textbf{Ours:} 10 FPS). 
\vspace{-0.5cm}} 

\label{fig:intro}
\end{figure}
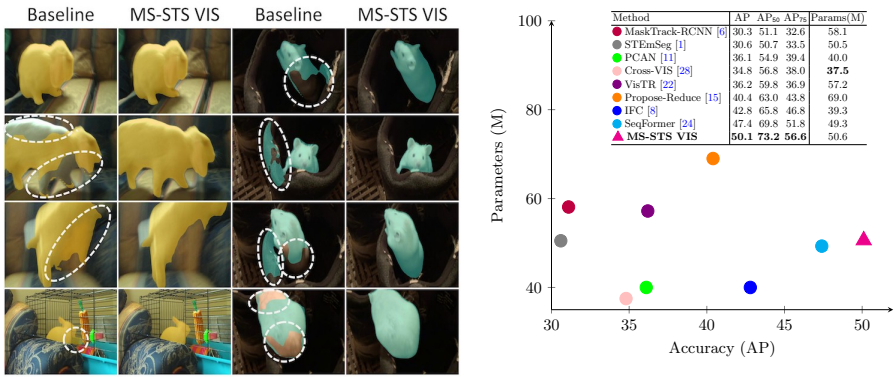

Recently, transformers~\cite{Attentionallyouneed} have shown promising results on several vision tasks, including VIS~\cite{vistr,SeqFormer}. The recent transformer-based VIS approaches~\cite{vistr,SeqFormer} are built on DETR~\cite{DETR} and Deformable DETR~\cite{Zhu_DeformableDETR_ICLR_2021} frameworks, utilizing an encoder-decoder architecture along with instance sequence matching and segmentation mechanisms to generate final video mask predictions. These approaches typically employ either single-scale spatio-temporal features~\cite{vistr} or per-frame multi-scale features~\cite{SeqFormer} during attention computations at the encoder and decoder. However, such an attention computation ignores the \textit{multi-scale spatio-temporal feature relationships}, which are crucial towards handling target appearance deformations due to real-world challenges such as, scale variation, change in aspect-ratio and fast motion in videos. 

In this work, we investigate the problem of designing an attention mechanism, within the transformer-based VIS framework, to effectively capture multi-scale spatio-temporal feature relationships in a video.
With this aim, we introduce a multi-scale spatio-temporal attention mechanism, which learns to aggregate the necessary attentions performed along the spatial and temporal axes without losing crucial information related to target appearance deformations in both the spatial and temporal axes. 
In addition to target appearance deformations, another major challenge in VIS is the accurate delineation of the target object in the presence of cluttered background. Surprisingly, existing transformer-based VIS approaches do not employ an explicit mechanism to enforce foreground-background (fg-bg) separability. Here, we introduce a loss formulation that improves fg-bg separability by emphasizing the fg regions in multi-scale spatio-temporal features while simultaneously suppressing the bg regions.

\noindent\textbf{Contributions:} We propose a transformer-based video instance segmentation framework, MS-STS VIS, with the following contributions.  
\begin{itemize}
    \item We propose a novel multi-scale spatio-temporal split (MS-STS) attention module in the transformer encoder for effectively capturing spatio-temporal feature relationships. Our MS-STS module first attends to features across frames at a given spatial scale via an intra-scale temporal attention block, and then progressively attends to neighboring spatial scales across frames via an inter-scale temporal attention block to obtain enriched feature representations. To further improve the video mask prediction, we introduce an attention block in the decoder that enhances the temporal consistency of the detected instances in different frames of a video.
    
    \item We introduce an auxiliary discriminator network during training to enhance the fg-bg separability within the multi-scale spatio-temporal feature space. Here, the discriminator network is trained to distinguish between ground-truth and predicted mask, while the encoder is learned to fool the discriminator by generating features that result in better mask predictions.

     \item Comprehensive experiments are performed on Youtube-VIS 2019 and 2021 datasets. Our proposed MS-STS VIS sets a new state-of-the-art on both datasets. When using the ResNet50 backbone, our MS-STS VIS outperforms all existing methods with an overall mask AP of 50.1\% on Youtube-VIS 2019 val. set, while being on par in terms of model size and speed compared to the state-of-the-art method (see Fig.~\ref{fig:intro} (right)). Specifically, the proposed MS-STS VIS achieves a significant performance improvement over the baseline, in case of target appearance deformations due to scale variation, aspect-ratio change and fast motion (see Fig.~\ref{fig:intro} (left)). 
\end{itemize}

\section{Baseline Framework} 
We base our approach on the recently introduced SeqFormer~\cite{SeqFormer}. SeqFormer is built on the Deformable DETR~\cite{Zhu_DeformableDETR_ICLR_2021} framework, which comprises a CNN backbone followed by a transformer encoder-decoder with deformable attention. We choose SeqFormer as our base framework to demonstrate the impact of our proposed contributions on a \textit{strong baseline}. Here, a video clip $\mathbf{x}$ consisting $T$ frames of spatial size $H^0\times W^0$ with a set of object instances is input to the backbone.  Latent feature maps for each frame are obtained from the backbone at $L$ multiple scales $2^{-l-2}$ with $1\leq l\leq L$ and passed through separate convolution filters for each scale. The output feature dimension for each of these convolution filters is set to $C$. The resulting multi-scale feature maps of each frame are then input to a transformer encoder comprising multi-scale deformable attention blocks. For each frame, the transformer encoder outputs multi-scale feature maps with the same size as its input. These encoder output features maps from each frame along with $n$ learnable instance query embeddings $\mathbf{I}^Q \in \mathbb{R}^C$ are then input to the transformer decoder comprising a series of self- and cross-attention blocks. The $n$ instance queries are further decomposed into $n$ box queries $\mathbf{B}^Q$ per-frame and are used to query the box features from the encoder feature maps of the corresponding frame. The learned box queries across $T$ frames are then aggregated temporally to obtain $n$ instance features $\mathbf{I}^O \in \mathbb{R}^C$. These instance features output by the decoder are then used for video instance mask prediction. We refer to~\cite{SeqFormer} for additional details.

\noindent \textbf{Limitation:} As discussed above, the aforementioned SeqFormer framework independently utilizes \textit{per-frame} multi-scale features during attention computations. As a result, it ignores the spatio-temporal feature relationships during attention computation that is crucial for the VIS problem. Different from SeqFormer that utilizes \textit{per-frame} spatial features at multiple scales, our approach performs \textit{multi-scale spatio-temporal attention} computation. Such a multi-scale spatio-temporal attention is especially desired in cases when the target object undergoes appearance deformations due to real-world challenges such as, scale variation, aspect-ratio change and fast motion in videos (see Fig.~\ref{fig:intro} and~\ref{fig:attention_maps}). Furthermore, distinct from the baseline SeqFormer, our approach employs an explicit mechanism to ensure accurate delineation of foreground objects from the cluttered background by enhancing fg-bg separability.

\section{Method} 

\begin{figure*}[t!]
    \centering
       \includegraphics[width=0.9\linewidth]{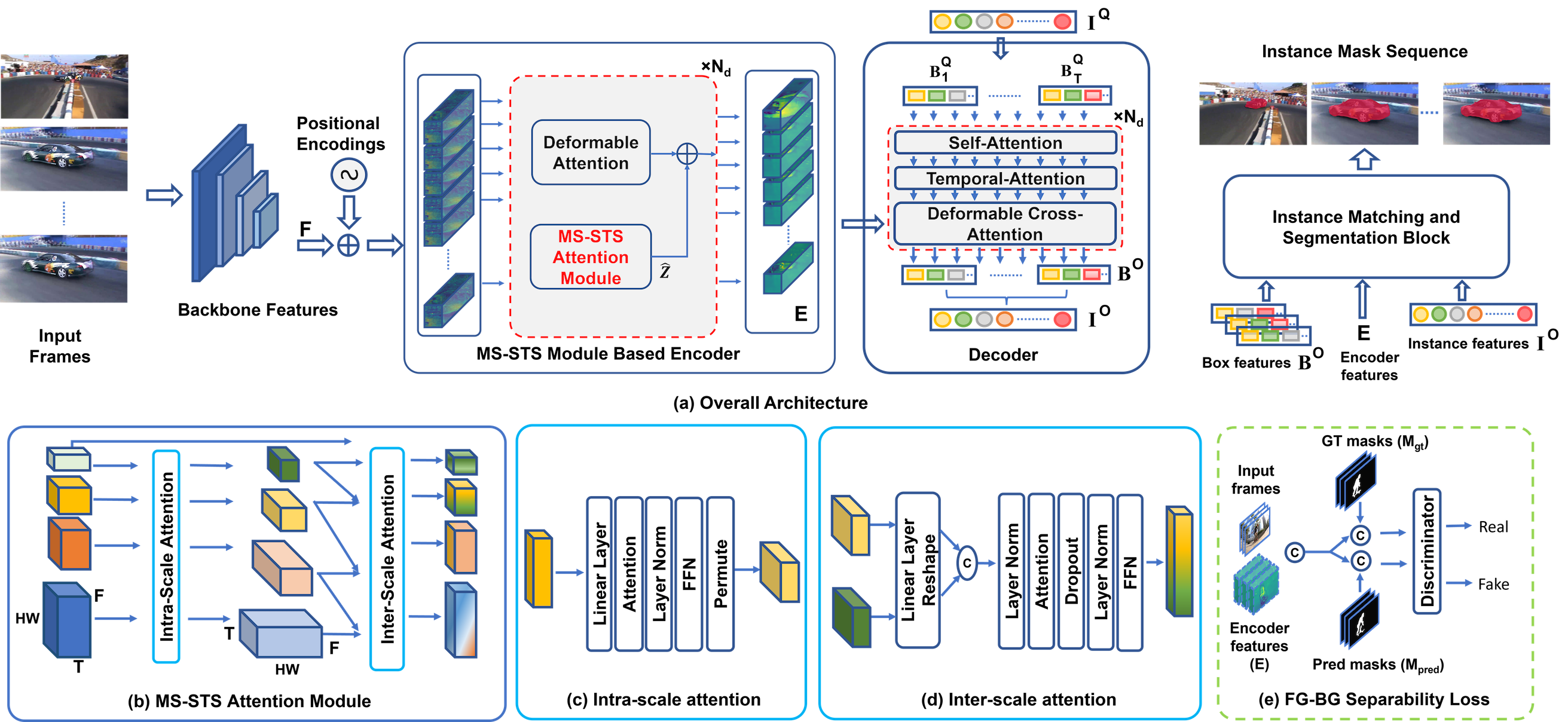}
    \vspace{-0.2cm}
    \caption{(a) Our MS-STS VIS architecture comprises a backbone, a transformer encoder-decoder and a instance matching and segmentation block. Here, our key contributions are: (i) a novel MS-STS attention module in the encoder to capture spatio-temporal feature relationships at multiple scales across frames, (ii) a temporal attention block in the decoder for enhancing the temporal consistency of the box queries and (iii) an adversarial loss for enhancing foreground-background (fg-bg) separability. Specifically, the MS-STS attention module (b) attends to the input backbone features by employing intra- and inter-scale temporal attentions to generate enriched features which are then fused with standard baseline features. While the intra-scale temporal attention block (c) enriches features across frames within a given scale, its inter-scale counterpart (d) progressively attends to multiple spatial scales across frames to obtain spatio-temporally enriched features. These enriched features are further improved using an adversarial loss (e) to enhance fg-bg separability. The resulting encoder features along with the temporally consistent instance features from the  decoder are used within the matching and segmentation block for the video instance mask prediction. 
  \vspace{-0.4cm}
  } 
    \label{fig:overall_architecture}
\end{figure*}

\subsection{Overall Architecture}
\label{subsubsec:overall_architecture}
Fig.~\ref{fig:overall_architecture}(a) shows the overall architecture of the proposed MS-STS VIS approach built on the baseline framework described above. Our MS-STS VIS comprises a backbone network, a transformer encoder-decoder and a sequence matching and segmentation block. The focus of our design is the introduction of a novel multi-scale spatio-temporal split (MS-STS) attention module (Fig.~\ref{fig:overall_architecture}(b)) in the transformer encoder to effectively capture spatio-temporal feature relationships at multiple scales across frames in a video. The MS-STS attention module comprises intra- and inter-scale temporal attention blocks (Fig.~\ref{fig:overall_architecture}(c) and (d)). The intra-scale block takes $C$-dimensional backbone features $\mathbf{F}$ as input and enriches features across frames in a video within a given scale, whereas the inter-scale block then progressively attends to multiple spatial scales across frames to generate spatio-temporally enriched features $C$-dimensional $\hat{\mathbf{Z}}$. These features $\hat{\mathbf{Z}}$ are fused in each encoder layer with the base features output by the standard deformable attention. While backbone features $\mathbf{F}$ are used as input to the first encoder layer, the subsequent layers utilize the outputs from the preceding layer as input. As a result, multi-scale spatio-temporally enriched features $C$-dimensional $\mathbf{E}$ are output by the encoder. 

Afterwards, these enriched features $\mathbf{E}$ from the encoder are input to the transformer decoder. To achieve temporal consistency among box queries from different frames, we introduce a temporal attention block within the transformer decoder. Next, the encoder features $\mathbf{E}$ along with the instance features $\mathbf{I}^O$ (aggregated temporally attended box queries) from the decoder are utilized within the instance matching and segmentation block to obtain the video instance mask prediction. To further improve the predicted video instance mask quality, we introduce an adversarial loss  during training to enhance foreground-background (fg-bg) separability. The adversarial loss (Fig.~\ref{fig:overall_architecture}(e)) strives to enhance the encoder features by discriminating between the predicted and ground-truth masks, utilizing the encoder features $\mathbf{E}$, the input frames $\mathbf{x}$ and the binary object mask $\mathbf{M}$. Next, we describe our MS-STS attention module-based encoder. 

\subsection{MS-STS Attention Module Based Encoder\label{sec:mr_sts_module}}

Within the proposed MS-STS VIS framework, we introduce a novel multi-scale spatio-temporal split (MS-STS) attention module in the transformer encoder to effectively capture spatio-temporal feature relationships at multiple scales across frames. To this end, the MS-STS module takes the backbone features as input and produces multi-scale spatio-temporally enriched features, which are then fused with the standard features within the base framework. The MS-STS module (see Fig.~\ref{fig:overall_architecture}(b)) comprises an intra-scale and an inter-scale attention block (see Fig.~\ref{fig:overall_architecture}(c) and (d)) described next.

\noindent\textbf{Intra-scale Temporal Attention Block:}
\label{subsubsection:intra_scale_Attention}
Given the backbone features as input, our intra-scale block independently attends to each scale (spatial resolution) temporally across frames (see Fig.~\ref{fig:overall_architecture}(c)). Let $\mathbf{z}_{s,t}^{l}$ be the feature at spatial scale $l$, position $s$ and frame $t$, where $s\in [0,S^l-1]$, $t\in [0,T-1]$, and $S^l=H^l \cdot W^l$. The intra-scale temporal attention block processes the features $\mathbf{Z}^{l} \in \mathbb{R}^{S^l \times T\times C}$ using intra-scale self-attention ($\text{SA}_{\text{intra}}$), layer normalization (LN), and MLP layers as:
\begin{equation}
    \mathbf{Y}^{l} = \mathrm{SA_{intra}}(\mathrm{LN}(\mathbf{Z}^{l})) + \mathbf{Z}^{l},  \qquad 
    \mathbf{\tilde{Z}}^{l} = \mathrm{MLP}(\mathrm{LN}(\mathbf{Y}^{l})) + \mathbf{Y}^{l},
\end{equation}    
\begin{equation}
    \text{where} \quad \mathrm{SA_{intra}}(\mathrm{LN}(\mathbf{z}^{l}_{s,t})) =  \sum_{\acute{t}=0}^{T-1} \mathrm{Softmax}\left(\frac{\mathbf{q}^{l}_{s,t}\cdot\mathbf{k}^{l}_{s,\acute{t}}}{\sqrt{C}}\right)\mathbf{v}^{l}_{s,\acute{t}}, 
\end{equation}
where $\mathbf{q}^{l}_{s,t},\mathbf{k}^{l}_{s,t}, \mathbf{v}^{l}_{s,t} \in \mathbb{R}^{D}$ are the query, key, and value vectors obtained from $\mathbf{z}_{s,t}^{l}$ (after LN) using the embedding matrices $\mathbf{W}_{q}, \mathbf{W}_{k},\mathbf{W}_{v} \in \mathbb{R}^{C \times C}$. The intra-scale temporal attention operates on each spatial scale $l$ across frames and produces temporally relevant intermediate features $\{\mathbf{\tilde{Z}}^{l}\}_{l=1}^{L}$.

\noindent\textbf{Inter-scale Temporal Attention Block:}
\label{subsubsection:inter_scale_Attention}
The inter-scale temporal attention block takes the intermediate features output by the intra-scale block and aims to learn similarities between the spatio-temporal features across two neighbouring spatial scales (see Fig.~\ref{fig:overall_architecture}(e)).
Let $\mathbf{\tilde{Z}}^{l} \in \mathbb{R}^{S^l \times T \times C}$ and $\mathbf{\tilde{Z}}^{l+1}$ be the intra-scale attended features at spatial scales $l$ and $l+1$. 
To compute inter-scale temporal attention between the two neighbouring scale features, we first upsample the lower resolution features by $\times$2 using bilinear interpolation, concatenate along the feature dimensions and project it into $C$ dimensions using $\mathbf{W}_{p} \in \mathbb{R}^{2C \times C}$ as:
\begin{equation}
    \mathbf{H}^{l} = \mathbf{W}_{p}(\mathrm{CONCAT}(\mathbf{\tilde{Z}}^{l}, \mathrm{UPSAMPLE}(\mathbf{\tilde{Z}}^{l+1})).
\end{equation}

\noindent We reshape $\mathbf{H}^{l} \in \mathbb{R}^{S^{l} \times T \times C}$ to $\mathbb{R}^{(S^{l} \cdot T) \times C}$ and then compute the joint spatio-temporal attention to obtain enriched features $\mathbf{\hat{Z}}^l$ given by

\begin{equation}
    \mathbf{Y}^{l} = \mathrm{SA_{inter}}(\mathrm{LN}(\mathbf{H}^{l})) + \mathbf{H}^{l}, \qquad \mathbf{\hat{Z}}^{l} = \mathrm{MLP}(\mathrm{LN}(\mathbf{Y}^{l})) + \mathbf{Y}^{l},
\end{equation}    

where the inter-scale self-attention ($\text{SA}_{\text{inter}}$) computation is given by
\begin{equation}
    \mathrm{SA_{inter}} = \sum_{\acute{s}=0}^{S-1} \sum_{\acute{t}=0}^{T-1} \mathrm{Softmax}\left(\frac{\mathbf{q}^{l}_{s,t}\cdot\mathbf{k}^{l}_{\acute{s},\acute{t}}}{\sqrt{C}}\right)\mathbf{v}^{l}_{\acute{s},\acute{t}}.
\end{equation}

\begin{figure*}[t!]
    \centering
       \includegraphics[width=1\linewidth]{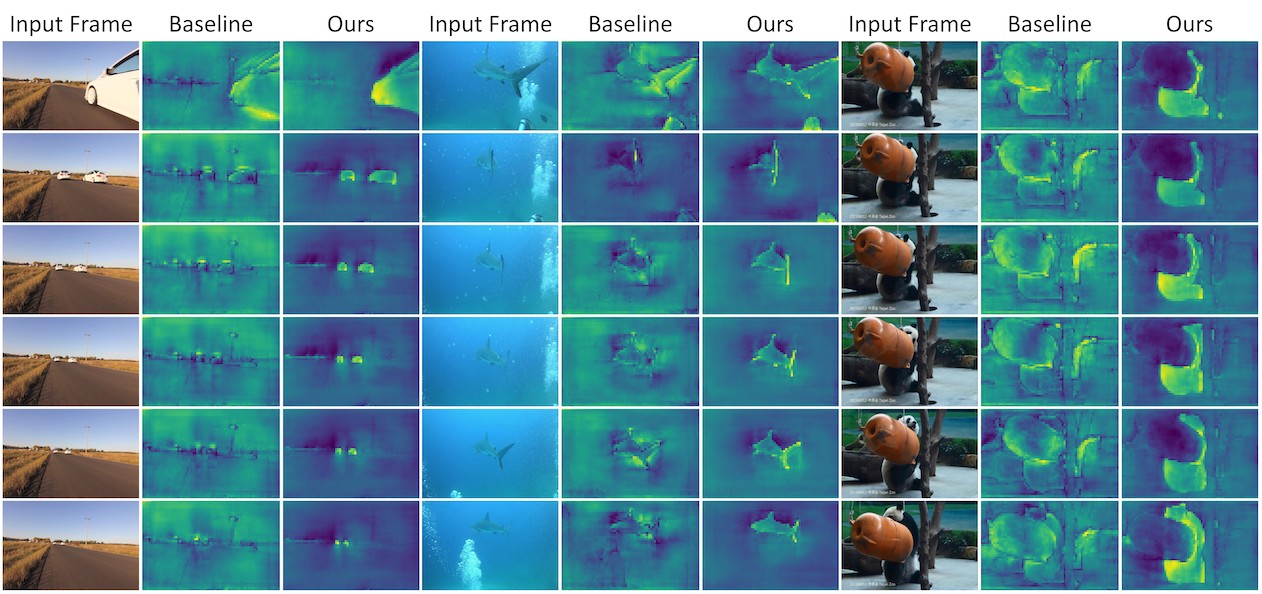}
    %   \vspace{-0.2cm}
    \caption{\label{fig:attention_maps}Example attention map visualizations obtained at the output of the baseline encoder and our MS-STS attention module-based encoder. Here, we show the comparison on example videos from Youtube-VIS 2019 val. set. The baseline struggles to accurately focus on the \textit{car} instance undergoing significant scale variation, where the size becomes extremely small in the later frames. Similarly, it fails to accurately focus on the \textit{shark} instance undergoing aspect-ratio change and the \textit{person} instance partially visible only in the first two frames of the middle video. In the last video, the baseline inaccurately highlights the irrelevant object (in orange) occluding the target \textit{panda} instance. 
     Our MS-STS attention module-based encoder, which strives to capture spatio-temporal feature relationships at multiple scales, successfully focuses on these challenging targets despite scale variations (\textit{car}), aspect-ratio change (\textit{shark}), partial visibility (\textit{person}) and appearance deformations due to occlusion (\textit{panda}). Best viewed zoomed in. 
    \vspace{-0.5cm}} 
   
\end{figure*}

To summarize, our MS-STS attention module utilizes an intra-scale temporal attention block to first attend to features across frames at a given spatial scale. It then employs an inter-scale temporal attention block to progressively attend to neighboring spatial scales across frames for obtaining enriched feature representations  $\mathbf{\hat{Z}}^l$. The resulting features $\mathbf{\hat{Z}}^l$ are fused with the standard baseline features through a convolution operation. Finally, the encoder outputs enriched multi-scale spatio-temporal features $\mathbf{E}$ after processing the input features through $N_d$ attention layers. Fig.~\ref{fig:attention_maps} shows example attention maps obtained at the output of our MS-STS attention module-based encoder. The attention maps are computed as the average activation strength across the $C$ features at a particular spatial position for the largest scale considered, \ie, $\nicefrac{H}{8} \times \nicefrac{W}{8}$. The attention maps are shown for example videos from Youtube-VIS 2019 val. set depicting target appearance deformations. We also compare our encoder output to that of the baseline encoder. Compared to the baseline, our MS-STS module-based encoder better focuses on the \textit{car} instance (left video) undergoing scale variations, the \textit{shark} instance (middle video) undergoing aspect-ratio change, the \textit{person} instance (middle video) partially visible only in the first two frames and the \textit{panda} instance (right video) exhibiting target deformations due to the irrelevant
object (in orange) occluding the target \textit{panda} instance.

\subsection{Enhancing Temporal Consistency in Decoder\label{sec:temporal_decoder}}

As discussed earlier, the transformer decoder in the base framework comprises a series (layers) of alternating self- and cross-attention blocks, operating on the box queries $\mathbf{B}^Q$ of individual frames. Although the video-level instance features $\mathbf{I}^O$ are obtained through an aggregation of the corresponding frame-level box features, the temporal consistency of the detected instances is hampered likely due to the per-frame attention computation of the box queries. To alleviate this issue, we introduce a temporal attention block in between the self- and cross-attention blocks of a decoder layer. Our temporal attention block attends to the sequence of box queries $\{\mathbf{B}_i^Q\}_{i=1}^{T}$ from $T$ frames and produces temporally consistent queries at its output. Such a temporal attention between the per-frame box queries of an instance enables information flow between the probable box locations across frames, thereby improving the temporal consistency of the video mask predictions. The resulting temporally-attended box queries are then utilized to query the multi-scale spatio-temporally enriched features $\mathbf{E}$, output by the encoder, for computing the box features $\mathbf{B}^O$ in a per-frame manner. Afterwards, these box features are aggregated to obtain video-level instance features $\mathbf{I}^O$. Finally, the resulting instance features along with the box features and multi-scale spatio-temporally enriched features $\mathbf{E}$ are input to an instance matching and segmentation block for generating the video mask predictions.

\subsection{Enhancing Foreground-Background Separability\label{sec:fgbg_separability}}
Both our MS-STS attention in encoder and temporal attention in decoder promote multi-scale spatio-temporal feature enrichment and temporal consistency, respectively, across frames. This helps to address the critical issue of appearance deformations in the target object due to challenges such as, scale variation, aspect-ratio change and fast motion. In addition to addressing target appearance deformations across frames, another common challenge in the VIS problem is the accurate delineation of foreground objects from the cluttered background. To this end, we introduce an adversarial loss during training of our MS-STS VIS framework for enhancing foreground-background (fg-bg) separability. To the best of our knowledge, we are the first to explore an adversarial loss within a transformer-based VIS framework for enhancing fg-bg separability. 

In our adversarial formulation, the objective is to discriminate between the ground-truth mask and the predicted mask output by our MS-STS VIS. With this objective, we introduce an auxiliary discriminator network during training. The discriminator takes the input frames along with the corresponding encoder features and binary masks as its input. Here, the binary mask $\mathbf{M}$ is obtained either from the ground-truth or predictions, such that all object instances (category-agnostic) within a frame are marked as foreground. While the discriminator $D$ attempts to distinguish between ground-truth and predicted binary masks ($\mathbf{M}_{gt}$ and $\mathbf{M}_{pred}$, respectively), the encoder learns to output enhanced features $\mathbf{E}$ such that the predicted masks $\mathbf{M}_{pred}$ are close to ground-truth $\mathbf{M}_{gt}$.
Let $\mathbf{F}_{gt} = \mathrm{CONCAT}(\mathbf{x},\mathbf{E},\mathbf{M}_{gt})$ and $\mathbf{F}_{pr} = \mathrm{CONCAT}(\mathbf{x},\mathbf{E},\mathbf{M}_{pred})$ denote the real and fake input, respectively, to the discriminator $D$. Similar to \cite{pix2pix}, the adversarial loss is then given by 
\begin{equation}
    \min_{Enc} \max_{D} \mathbb{E}[\log D(\mathbf{F}_{gt})] + \mathbb{E}[\log (1 - D(\mathbf{F}_{pr}))] + \lambda_1 \mathbb{E}[||D(\mathbf{F}_{gt})-D(\mathbf{F}_{pr})||_1].
\end{equation}
Since the mask prediction depends on the quality of the encoder features that are decoded by the queries, we treat our encoder $Enc$ as a generator in our adversarial formulation above. As a result, the encoder learns to better delineate foreground and background regions leading to improved video instance mask prediction. Note that the discriminator network is utilized only during  training. 

\section{Experiments} 

\subsection{Experimental Setup}
\noindent \textbf{Datasets:}
The \textit{YouTube-VIS 2019} \cite{YouTube-VIS-2019} dataset contains 2883 high-quality videos with 131$K$ annotated object instances belonging to 40 different categories. 
The \textit{YouTube-VIS 2021} \cite{YouTube-VIS-2021} dataset contains 3,859 high-quality videos with 232$K$ annotated object instances belonging to 40 different improved categories. 
YouTube-VIS 2021\cite{YouTube-VIS-2021} has a 40-category label set by merging \textit{eagle} and \textit{owl} into \textit{bird}, \textit{ape} into \textit{monkey}, deleting \textit{hands}, and adding \textit{flying disc}, \textit{squirrel} and \textit{whale}, maintaining the same number of categories as YouTube-VIS 2019\cite{YouTube-VIS-2019} set.\\
\noindent \textbf{Evaluation Metrics:} We follow the standard protocol, where the evaluation metrics, Average Precision (AP) and Average Recall (AR), are adapted from image instance segmentation with the video Intersection over Union (IoU) of the mask sequences as the threshold. 

\noindent \textbf{Implementation Details:}
We choose ResNet-50~\cite{ResNet} as the default backbone, unless otherwise specified, for our MS-STS VIS framework. Outputs from conv${}_3$, conv${}_4$ and conv${}_5$ of the Resnet backbone are utilized to obtain multi-scale feature inputs to our transformer encoder-decoder, as in~\cite{Zhu_DeformableDETR_ICLR_2021}. Both encoder and decoder layers are set to $N_d{=}6$. The feature dimension $C$ is set to 256, while the number of instance queries is set to 300 and length of video clip $T{=}5$, as in~\cite{SeqFormer}. We use the AdamW optimizer with a base learning rate (LR) of 2$\times10^{-4}$, ($\beta_1$, $\beta_2) {=} (0.9, 0.999)$ and a weight decay of $10^{-4}$. LR of linear projections of deformable attention modules and the backbone are scaled by $0.1$. The model is first pretrained on COCO~\cite{MSCOCO} for 24 epochs with a batch size of 2. Similar to~\cite{SeqFormer}, the pretrained weights are then used to train the model on Youtube-VIS and COCO dataset for 12 epochs with $T{=}5$ and batch size set to 2. The LR is scaled by a factor of $0.1$ at $4^{th}$ and $10^{th}$ epochs. The framework is trained on 8 Nvidia V100 GPUs using PyTorch-1.9~\cite{pytorch}. 

\begin{table*}[t!]
\centering
\caption{State-of-the-art comparison on \textbf{YouTube-VIS 2019} $\mathtt{val}$ set. Our MR-STS VIS consistently outperforms the state-of-the-art results reported in literature. When using the ResNet-50 backbone, MS-STS VIS achieves overall mask AP score of 50.1\% with an absolute gain of 2.7\% over the best existing SeqFormer, while being comparable in terms of model size and speed (SeqFormer: 11 FPS \vs{} Ours: 10 FPS). Similarly, when using the ResNet-101 backbone, our MS-STS VIS achieves overall mask AP of 51.1\%. Further, MS-STS VIS achieves the best accuracy reported on this dataset with a mask AP of 61.0\% and outperforms SeqFormer with an absolute gain of 1.7\%, using the same Swin-L backbone.\vspace{-0cm}
 }
 \setlength{\tabcolsep}{6pt}
 \adjustbox{width=\textwidth}{
 \begin{tabular}{l|l|c|c|ccccc}
 \toprule[0.15em]
  \textbf{Method}
  & \textbf{Venue}
  & \textbf{Backbone}
  & \textbf{Type}
  & \textbf{AP} 
  & \textbf{AP}$_{\mathtt{50}}$ 
  & \textbf{AP}$_{\mathtt{75}}$ 
  & \textbf{AR}$_{\mathtt{1}}$ 
  & \textbf{AR}$_{\mathtt{10}}$ \\
  \toprule[0.15em]
  IoUTracker+ \cite{YouTube-VIS-2019} & ICCV 2019 & ResNet-$50$ & - & 23.6 & 39.2 & 25.5 & 26.2 & 30.9 \\
  OSMN \cite{OSMN} & CVPR 2018 & ResNet-$50$ & Two-Stage & 27.5 & 45.1 & 29.1 & 28.6 & 33.1 \\
  DeepSORT \cite{DeepSort} & ICIP 2017 & ResNet-$50$ & Two-stage & 26.1 & 42.9 & 26.1 & 27.8 & 31.3 \\
  FEELVOS \cite{FeelVOS} & CVPR 2019 & ResNet-$50$ &  Two-stage & 26.9 & 42.0 & 29.7 & 29.9 & 33.4 \\
  SeqTracker \cite{YouTube-VIS-2019} & ICCV 2019 & ResNet-$50$ & - & 27.5 & 45.7 & 28.7 & 29.7 & 32.5 \\
  MaskTrack R-CNN \cite{YouTube-VIS-2019} & ICCV 2019 & ResNet-$50$ & Two-stage & 30.3 & 51.1 & 32.6 & 31.0 & 35.5 \\
  MaskProp \cite{MaskPropagation} & CVPR 2020 & ResNet-$50$ & - & 40.0 & - & 42.9 & - & - \\
  SipMask-VIS \cite{SipMask} & ECCV 2020 & ResNet-$50$ & One-stage & 32.5 & 53.0 & 33.3 & 33.5 & 38.9 \\
  SipMask-VIS \cite{SipMask} & ECCV 2020 & ResNet-$50$  & One-stage & 33.7 & 54.1 & 35.8 & 35.4 & 40.1 \\
  STEm-Seg \cite{STEmSeg} & ECCV 2020 & ResNet-$50$  & - & 30.6 & 50.7 & 33.5 & 31.6 & 37.1 \\
  Johnander \etal \cite{VIS_RGNN} & GCPR 2021 & ResNet-$50$  & - & 35.3 & - & - & - & - \\
  CompFeat \cite{CompFeat} & AAAI 2021 & ResNet-$50$ & - & 35.3 & 56.0 & 38.6 & 33.1 & 40.3 \\
  CrossVIS\cite{CROSS-VIS} &  ICCV 2021 & ResNet-$50$  & One-stage & 36.3 & 56.8 & 38.9 & 35.6 & 40.7 \\
  PCAN \cite{PCAN} & NeurIPS 2021 & ResNet-$50$ & One-stage & 36.1 & 54.9 & 39.4 & 36.3 & 41.6 \\
  VisTR \cite{vistr} & CVPR 2021 & ResNet-$50$ & Transformer & 35.6 & 56.8 & 37.0 & 35.2 & 40.2 \\
  SeqFormer \cite{SeqFormer} & Arxiv 2021 & ResNet-$50$ & Transformer & 47.4 & 69.8 & 51.8 & 45.5 & 54.8 \\
  \textbf{MS-STS VIS (Ours)} &  & ResNet-$50$ & Transformer  & \textbf{50.1} & \textbf{73.2} & \textbf{56.6} & \textbf{46.1} & \textbf{57.7} \\
  \midrule
  MaskTrack R-CNN \cite{YouTube-VIS-2019}& ICCV 2019 & ResNet-$101$ & Two-stage & 31.9 & 53.7 & 32.3 & 32.5 & 37.7 \\
  MaskProp \cite{MaskPropagation} & CVPR 2020 & ResNet-$101$  & - & 42.5 & - & 45.6 & - & - \\
  STEm-Seg \cite{STEmSeg} 
  & ECCV, 2020
  & ResNet-$101$ 
  & -
  & 34.6 & 55.8 & 37.9 & 34.4 & 41.6 \\
  CrossVIS \cite{CROSS-VIS} & ICCV 2021 & ResNet-$101$ & One-stage & 36.6 & 57.3 & 39.7 & 36.0 & 42.0 \\
  PCAN \cite{PCAN} & NeurIPS 2021 & ResNet-$101$ & One-stage & 37.6 & 57.2 & 41.3 & 37.2 & 43.9 \\
  VisTR \cite{vistr} & CVPR 2021 & ResNet-$101$ & Transformer & 38.6 & 61.3 & 42.3 & 37.6 & 44.2 \\
  SeqFormer \cite{SeqFormer} & Arxiv 2021 & ResNet-$101$ & Transformer & 49.0 & 71.1 & 55.7 & 46.8 & 56.9 \\
  \textbf{MS-STS VIS (Ours)} &  & ResNet-$101$ & Transformer  & \textbf{51.1} & \textbf{73.2} & \textbf{59.0} & \textbf{48.3} & \textbf{58.7} \\
  \midrule
  SeqFormer \cite{SeqFormer} & Arxiv 2021 & Swin-L & Transformer & 59.3 & 82.1 & 66.6 & 51.7 & 64.4 \\
  \textbf{MS-STS VIS (Ours)} &  & Swin-L & Transformer  & \textbf{61.0} & \textbf{85.2} & \textbf{68.6} & \textbf{54.7} & \textbf{66.4} \\
 \bottomrule[0.1em]
 \end{tabular} 
 }
 \vspace{-0.4cm}
 \label{tab:sota_valset_2019}
\end{table*}

\subsection{State-of-the-art Comparison} 
Tab.~\ref{tab:sota_valset_2019} presents the state-of-the-art comparison on the YouTube-VIS 2019 val. set. When using the ResNet-50 backbone, the recent one-stage PCAN~\cite{PCAN} and CrossVIS~\cite{CROSS-VIS} approaches achieve an overall mask accuracy (AP) of 36.1\% and 36.3\%, respectively. With the same ResNet-50 backbone, the first transformer-based VIS approach, VisTR~\cite{vistr}, built on DETR framework achieves an overall mask AP of 35.6\%. Among existing methods, the recently introduced SeqFormer~\cite{SeqFormer} based on Deformable DETR framework achieves the best overall accuracy with a mask AP of 47.4\%. Our proposed MS-STS VIS approach outperforms SeqFormer~\cite{SeqFormer} by achieving an overall mask AP of 50.1\%, using the same ResNet-50 backbone. Specifically, our MS-STS VIS provides an absolute gain of 4.8\% at a higher overlap threshold of AP$_{\mathtt{75}}$ over SeqFormer. Similarly, our MS-STS VIS consistently outperforms SeqFormer with an overall mask AP of 51.1\%, when using the ResNet-101 backbone. Finally, when using the recent Swin Transformer backbone, the proposed MS-STS VIS achieves the best accuracy reported in literature with an overall mask AP of 61.0\%.

\begin{table}[t!]
\centering
\caption{\label{tab:youtube-vis-2021}State-of-the-art comparison on \textbf{YouTube-VIS 2021} $\mathtt{val}$ set. All results are reported using the same ResNet-50 backbone. Our MS-STS VIS achieves state-of-the-art results with an overall mask AP of 42.2\% and an absolute gain of 2.8\% over the best existing SeqFormer at a higher overlap threshold of AP$_{\mathtt{75}}$. } 
 \setlength{\tabcolsep}{10pt}
 \adjustbox{width=0.7\textwidth}{
 \begin{tabular}{l|ccccc}
  \toprule[0.15em]
  \textbf{Method} 
  & \textbf{AP} 
  & \textbf{AP}$_{\mathtt{50}}$ 
  & \textbf{AP}$_{\mathtt{75}}$ 
  & \textbf{AR}$_{\mathtt{1}}$ 
  & \textbf{AR}$_{\mathtt{10}}$ 
  \\
 \toprule[0.15em]
  MaskTrack R-CNN\cite{YouTube-VIS-2021} & $28.6$ & $48.9$ & $29.6$ & $26.5$ & $33.8$ \\
  SipMask-VIS\cite{SipMask} & $31.7$ & $52.5$ & $34.0$ & $30.8$ & $37.8$ \\
  VisTR\cite{vistr} & $31.8$ & $51.7$ & $34.5$ & $29.7$ & $36.9$ \\
  CrossVIS\cite{CROSS-VIS} & $34.2$ & $54.4$ & $37.9$ & $30.4$ & $38.2$ \\
  IFC\cite{ifc} & $36.6$ & $57.9$ & $39.9$ & $-$ & $-$ \\
  SeqFormer\cite{SeqFormer} & $40.5$ & $62.4$ & $43.7$ & $36.1$ & $48.1$ \\
  \textbf{MS-STS VIS (Ours)} & \textbf{42.2} & \textbf{63.7} & \textbf{46.5} & \textbf{41.7} & \textbf{51.1} \\
 \bottomrule[0.15em]
 \end{tabular}
 }
 \vspace{-0.4cm}
\end{table}

Tab.~\ref{tab:youtube-vis-2021} reports the state-of-the-art comparison on the YouTube-VIS 2021 val. set. Among existing methods, CrossVIS~\cite{CROSS-VIS} and IFC~\cite{ifc} achieve overall mask AP scores of 34.2\% and 36.6\%, respectively. SeqFormer \cite{SeqFormer} obtains an overall mask AP of 40.5\%. Our MS-STS VIS sets a new state-of-the-art with an overall mask AP of 42.2\%. Specifically, MS-STS VIS provides an absolute gain of 2.8\% over SeqFormer at higher overlap threshold of AP$_{\mathtt{75}}$, when using the same ResNet-50 backbone. 

\begin{table}[t!]
\centering
\caption{\label{tab: ablation1}\textbf{On the left:} Impact of our contributions when progressively integrating them into the baseline on the Youtube-VIS 2019 val. set. We observe a consistent performance improvement due to the integration of our proposed contributions. Our final MS-STS VIS (row 4) achieves an absolute gain of 3.7\% over the baseline. \textbf{On the right:} Attribute-based performance comparison between the baseline and our  MS-STS VIS on the custom set comprising 706 videos in total. Here, we present the comparison on fast motion, target size change and aspect-ratio change attributes. Our MS-STS VIS achieves consistent improvement in performance over the baseline on all attributes. See Sec.~\ref{sec:ablation} for details.\vspace{-0cm}
 }
\resizebox{\textwidth}{!}{
  \begin{tabular}{l|l}
   \begin{tabular}{l|c}
         \toprule[0.1em]
            \textbf{Model}
          & \textbf{AP} 
          \\
          \toprule[0.1em]
           Baseline & $46.4$ \\
           Baseline + MS-STS & $48.4$ \\
           Baseline + MS-STS + T-Dec & $49.1$ \\
           Baseline + MS-STS + T-Dec + FG-BG Loss & \textbf{50.1} \\
          \bottomrule[0.1em]
          \end{tabular}
  &
   \begin{tabular}{|l@{\,\;\,}|c@{\,\;\,}c@{\,\;\,}c}
          \toprule[0.1em]
            \textbf{Attribute Type} & \textbf{\#Videos} & \textbf{Baseline} &  \textbf{MS-STS VIS} \\
          \toprule[0.1em]
           Custom Set & $ 706 $ & $ 49.5 $ & \textbf{53.7} \\
           Fast Motion & $ 605 $ & $ 48.9 $ & \textbf{54.2} \\
           Target Size Change & $ 706 $ & $ 49.5 $ & \textbf{53.7} \\
           Aspect-Ratio Change & $ 687 $ & $ 49.2 $ & \textbf{53.9} \\
          \bottomrule[0.1em]
         \end{tabular}
    \end{tabular}}
  \vspace{-0.2cm}
\end{table}

\subsection{Ablation Study\label{sec:ablation}} 
Here, we first evaluate the merits of our three proposed contributions: MS-STS attention module-based encoder (Sec.~\ref{sec:mr_sts_module}), temporal attention in the decoder (Sec.~\ref{sec:temporal_decoder}) and the adversarial loss for enhancing fg-bag separability (Sec.~\ref{sec:fgbg_separability}). Tab.~\ref{tab: ablation1} (left) shows the baseline comparison on the YouTube-VIS 2019 val. set. All results reported in Tab.~\ref{tab: ablation1} (left) are obtained using the same ResNet50 backbone. As discussed earlier, our MS-STS VIS employs SeqFormer as its base framework.  We train the baseline SeqFormer (denoted here as Baseline) using the official implementation and achieve an overall mask AP score of 46.4\%. The introduction of our MS-STS attention module-based encoder within the baseline (referred as Baseline + MS-STS) significantly improves the overall performance to 48.4\% with an absolute gain of 2.0\%. The overall performance is further improved to 49.1\% with the integration of the temporal attention in the decoder (denoted as Baseline + MS-STS + T-Dec). Finally, the introduction of the adversarial loss during the training for enhancing the fg-bg separability provides an absolute gain of 1.0\% (denoted as Baseline + MS-STS + T-Dec + FG-BG Loss). Our final MS-STS VIS achieves an absolute gain of 3.7\% over the baseline. 

We further analyze the performance of our method (MS-STS VIS) under three specific challenging scenarios: fast motion, target size change (scale variation) and aspect-ratio change. To this end, we classified the videos into three categories: (i) fast motion of the object, (ii) object size changes, and (iii) aspect-ratio changes. In particular, we follow the well-established VOT-2015 benchmark \cite{VOT-2015} to label a particular video from above categories as follows:
  (i)  
  \textbf{fast motion:} if object center in current frame moves by at least 30\% of its size in previous frame.
  (ii)	
  \textbf{change in object size:} if the ratio of the maximum size to the minimum size of an object in the video is greater than 1.5.	
  (iii) 
  \textbf{change in aspect-ratio:} if the ratio of the maximum to the minimum aspect (width/height) of the bounding box enclosing an object in the video is greater than 1.5.
  
  Based on the aforementioned criteria, we first select the newly added videos in Youtube-VIS 2021 training set while retaining the same set of classes as in Youtube-VIS 2019 dataset. We refer these selected videos as custom set and classify them into the aforementioned attributes. To evaluate the performance of our MS-STS VIS and the baseline, we use the models trained on Youtube-VIS 2019 training set with the same ResNet-50 backbone. Note that we ensure that there is no overlap between the videos in Youtube-VIS 2019 training set and our custom set (only comprising the newly added videos from the Youtube-VIS 2021 training set). Tab.~\ref{tab: ablation1} (right) shows the comparison between the baseline SeqFormer and our MS-STS VIS. On the entire custom set, our MS-STS VIS obtains significantly improved performance over the baseline. Specifically, MS-STS VIS achieves absolute gains of 5.3\%, 4.2\% and 4.7\% over the baseline on fast motion, target size change and aspect-ratio change attributes, respectively. 

\subsection{Qualitative Analysis\label{sec:qual_res}}
\begin{figure*}[t!]
    \centering
       \includegraphics[width=1\linewidth]{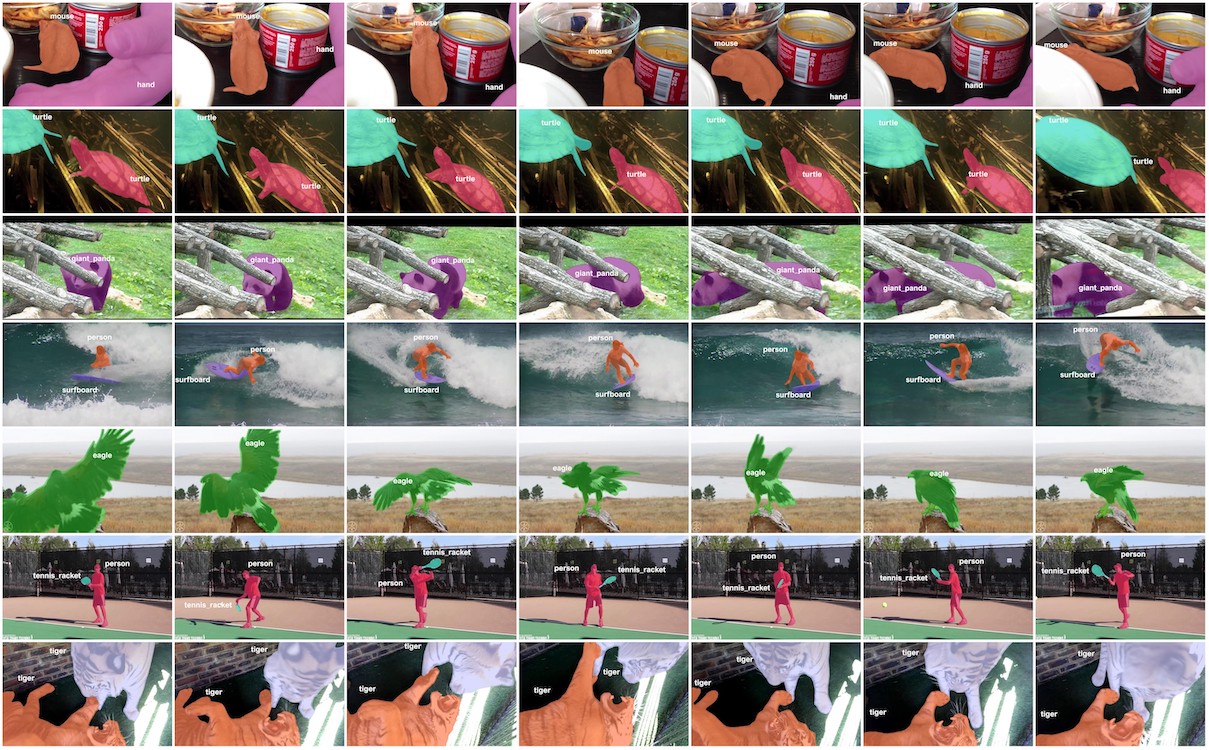}\vspace{-0.3cm}
    \caption{Qualitative results obtained by our MS-STS VIS framework on seven example videos in the Youtube-VIS 2019 val set. Our MS-STS VIS achieves promising video mask prediction in various challenging scenarios including, fast motion (\textit{eagle} in row 5), scale change (\textit{hand} in row 1), aspect-ratio change (\textit{panda} in row 3, \textit{person} in row 4), multiple instances of same class (\textit{tiger} in row 7).\vspace{-0.5cm}} 
    \label{fig:qual_results}
\end{figure*}

\begin{figure*}[t!]
    \centering
    \includegraphics[width=1\linewidth]{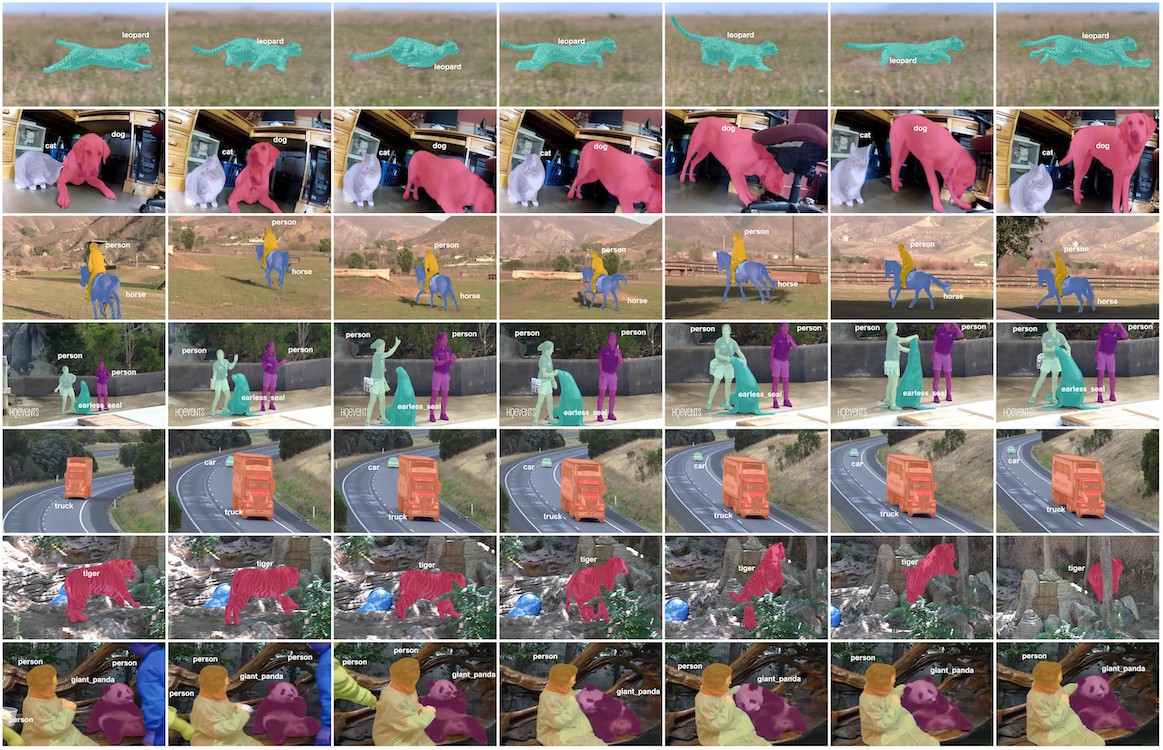}\vspace{-0.3cm}
    \caption{Qualitative results on seven example videos in the Youtube-VIS 2021 val set. Our MS-STS VIS achieves favorable video mask prediction in various scenarios involving target appearance deformations: fast motion (\textit{leopard} in row 1), scale variation (\textit{person} and \textit{seal} in row 4), aspect-ratio change (\textit{dog} in row 2, \textit{tiger} in row 6).\vspace{-0.4cm}}
    \label{fig:qual_results_21}
\end{figure*}

Fig.~\ref{fig:qual_results} and~\ref{fig:qual_results_21} show qualitative results obtained by our MS-STS VIS framework on example videos from the Youtube-VIS 2019 val. and 2021 val. sets, respectively. We observe our MS-STS VIS framework to obtain promising video mask prediction in various challenging scenarios involving target appearance deformations due to fast motion, aspect-ratio change and scale variation. \Eg, in Fig.~\ref{fig:qual_results}, video masks are predicted accurately for \textit{hand} in row 1 (scale change), \textit{eagle} in row 5 (fast motion, aspect-ratio change, scale variation), \textit{panda} in row 3 (aspect-ratio change), \etc. Similarly, we observe promising video masks predictions for \textit{leopard} in row 1 (fast motion), \textit{person} in row 3 (scale variation), \textit{dog} in row 2 (aspect-ratio change, scale variation) in Fig.~\ref{fig:qual_results_21}. These results show the efficacy of our MS-STS VIS framework under different challenges for the task of video instance segmentation. 

\section{Related Work}
\label{sec:relatedwork}
\noindent \textbf{Two-stage VIS:} Several VIS methods \cite{MaskPropagation,YouTube-VIS-2019,VAEVIS} adapt the two-stage pipeline, such as Mask R-CNN \cite{MaskRCNN} by introducing an additional tracking branch for target association. The work of \cite{YouTube-VIS-2019} introduces MaskTrack R-CNN that jointly performs detection, segmentation and tracking. Bertasius \etal{} \cite{MaskPropagation} utilize a branch in Mask R-CNN to propagate frame-level instance masks based on deformable convolutions from each frame to other video frames within a temporal neighborhood. The work of \cite{VAEVIS} introduces a modified variational autoencoder (VAE) on top of Mask R-CNN for instance-level video segmentation and tracking.\\
\noindent \textbf{Single-stage VIS:} Several works \cite{SipMask,STEmSeg,PCAN,STMask,SGNet} adapt the one-stage pipeline, such as FCOS detector \cite{FCOS}, where a linear combination of mask bases are directly predicted as final segmentation. SipMask \cite{SipMask} introduces a spatial information preservation module for real-time VIS. The work of \cite{STEmSeg} introduces an approach where a short 3D convolutional spatio-temporal volume is adopted to learn pixel-level embedding by posing segmentation as a bottom-up grouping. The work of \cite{PCAN} proposes to refine a space-time memory into a set of instance and frame-level prototypes, followed by an attention scheme. 

\noindent\textbf{Transformer-based VIS:} 
Wang \etal{} \cite{vistr} introduce a transformer-based encoder-decoder architecture, named VisTR, that formulates VIS as a direct end-to-end parallel sequence prediction task. 
In the encoder, VisTR utilizes a single-scale attention that computes similarities between all pairs of features from multiple spatial locations of a low-resolution feature map, across frames. The work of \cite{ifc} introduces inter-frame communication transformers, where memory tokens are used to communicate between frames. The recent SeqFormer \cite{SeqFormer}, built on Deformable DETR framework \cite{Zhu_DeformableDETR_ICLR_2021},  utilizes per-frame multi-scale features during attention computations. While demonstrating promising results, SeqFormer struggles in case of target deformations likely due to not explicitly capturing the spatio-temporal feature relationships during attention computation. To address these issues, we proposed a framework comprising an encoder that captures multi-scale spatio-temporal feature relationships. We also introduced an attention block in decoder to enhance temporal consistency of detected instance in different frames and an adversarial loss during training that ensures better fg-bg separability within  multi-scale spatio-temporal feature space. 

\section{Conclusions}
We proposed a transformer-based video instance segmentation framework, named MS-STS VIS, which comprises a novel multi-scale spatio-temporal split attention (MS-STS) module to effectively capture spatio-temporal feature relationships at multiple scales across frames in a video.  We further introduced an auxiliary discriminator network during training that strives to enhance fg-bg separability within the multi-scale spatio-temporal feature space. Our MS-STS VIS  specifically tackles target appearance deformations due to real-world challenges such as, scale variation, aspect-ratio change and fast motion in videos. Our extensive experiments on two datasets reveal the benefits of the proposed contributions, achieving state-of-the-art performance on both benchmarks. 

\section*{Acknowledgements}
This work was partially supported by VR starting grant (2016-05543), the Wallenberg AI, Autonomous Systems and Software Program (WASP), by the Swedish Research Council through a grant for the project Algebraically Constrained Convolutional Networks for Sparse Image Data (2018-04673), and the strategic research environment ELLIIT, in addition to the compute support provided at the Swedish National Infrastructure for Computing (SNIC), partially funded by the Swedish Research Council through grant agreement no. 2018-05973.

\bibliographystyle{splncs04}
\bibliography{egbib}
\end{document}